\documentclass[letterpaper, 10 pt, journal, twoside]{IEEEtran}
\usepackage[english]{babel}
\usepackage{amsmath} 
\usepackage{amssymb}  
\usepackage{graphicx}
\usepackage{subfigure}
\usepackage{multirow}
\usepackage{amsfonts}
\usepackage{hyperref}
\usepackage{tabularx}
\usepackage[noend]{algpseudocode}
\usepackage{algorithm}
\usepackage{bm}
\usepackage{enumerate}
\usepackage{verbatim}
\usepackage{float}
\usepackage{siunitx}
\usepackage{mdframed}
\usepackage{adjustbox}
\usepackage{cleveref}
\usepackage{authblk}
\usepackage{color}
\usepackage{soul}
\usepackage{cite}
\usepackage{booktabs}

\usepackage{xcolor}
\definecolor{myPurple}{RGB}{150, 130, 205}
\definecolor{myYellow}{RGB}{204, 170, 0} 

\ifCLASSINFOpdf
\else
\fi
\begin{document}
%
\title{SonicBoom: Contact Localization Using\\ Array of Microphones}

%

\author{
Moonyoung Lee, Uksang Yoo, Jean Oh, Jeffrey Ichnowski, George Kantor, Oliver Kroemer%

\thanks{$^{1}$ Moonyoung Lee, Uksang Yoo, Jean Oh, Jeffery Ichnowski, George Kantor, Oliver Kroemer are with the Robotics Institute, Carnegie Mellon University, Pittsburgh 15206 USA
        {\tt\footnotesize \{moonyoul, uyoo, hyaejino, jichnows, kantor,  okroemer\}@cs.cmu.edu}}%
}





\maketitle

\definecolor{navy}{RGB}{0,0,128}
\definecolor{darkgreen}{rgb}{0,0.5,0}

\newcommand{\TODO}[1]{{\color{blue}{#1}}}


\begin{abstract}

 In cluttered environments where visual sensors encounter heavy occlusion, such as in agricultural settings, tactile signals can provide crucial spatial information for the robot to locate rigid objects and maneuver around them. We introduce SonicBoom, a holistic hardware and learning pipeline that enables contact localization through an array of contact microphones. While conventional sound source localization methods effectively triangulate sources in air, localization through solid media with irregular geometry and structure presents challenges that are difficult to model analytically. We address this challenge through a feature-engineering and learning-based approach, autonomously collecting 18,000 robot interaction-sound pairs to learn a mapping between acoustic signals and collision locations on the robot end-effector link. By leveraging relative features between microphones, SonicBoom achieves localization errors of 0.43cm for in-distribution interactions and maintains robust performance of 2.22cm error even with novel objects and contact conditions. We demonstrate the system's practical utility through haptic mapping of occluded branches in mock canopy settings, showing that acoustic-based sensing can enable reliable robot navigation in visually challenging environments. Our research platform is open-sourced, with additional information available at  \textcolor{navy}{\url{https://iamlab-cmu.github.io/sonicboom}}.

\end{abstract}


\IEEEpeerreviewmaketitle


\section{Introduction}

\begin{figure}[t]
    \centering
    \includegraphics[width=1.0\linewidth]{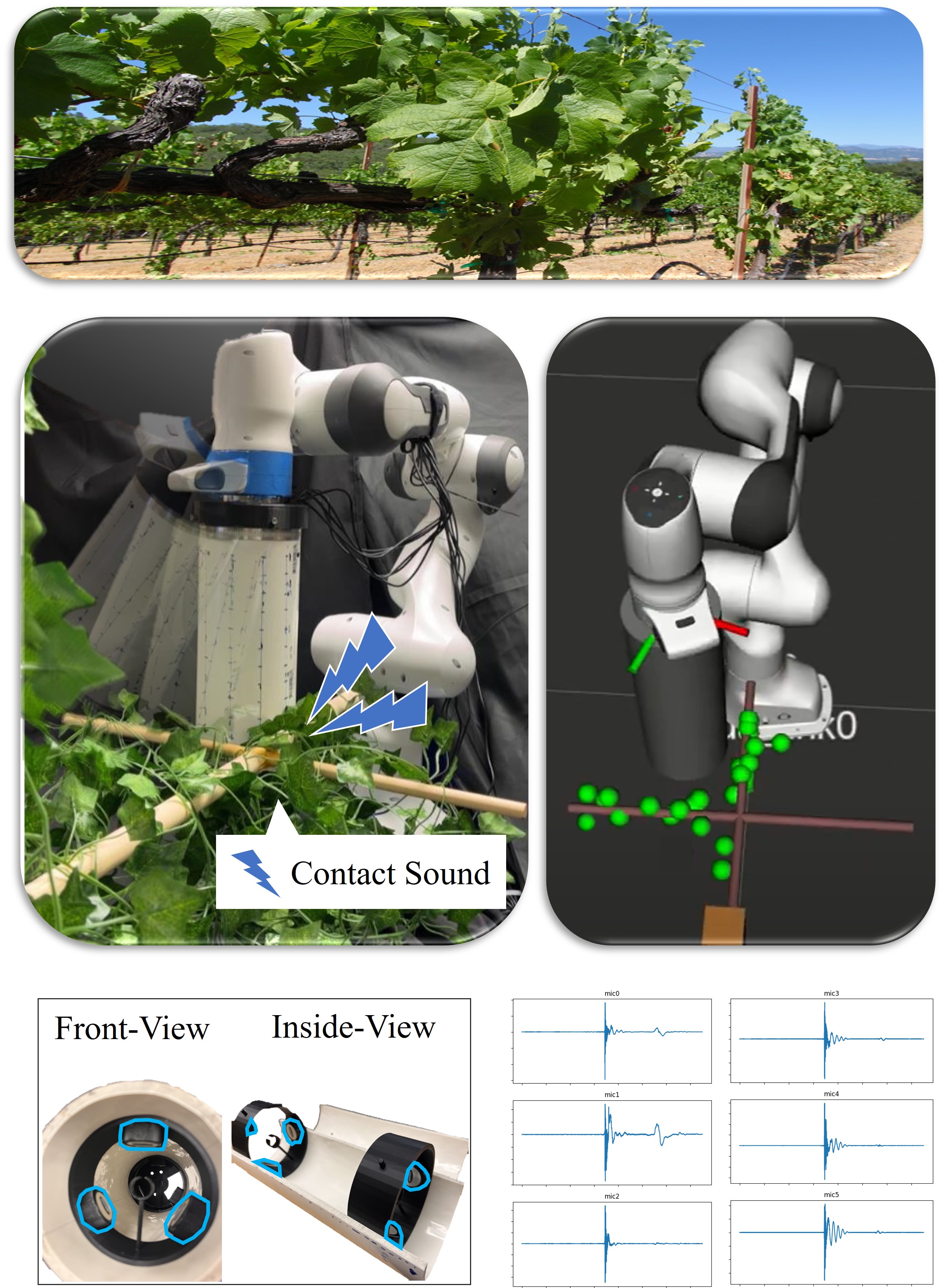}
    \setlength{\unitlength}{1cm}
    \begin{picture}(0,0)
    \put(.2, 9.6){\textcolor{black}{\small (a)}}
    \put(-2, 3.3){\textcolor{black}{\small (b)}}
    \put(2, 3.3){\textcolor{black}{\small (c)}} 
    \put(-2, 0.1){\textcolor{black}{\small (d)}}
    \put(2, 0.1){\textcolor{black}{\small (e)}}
    
    \end{picture}
    \vspace{-5pt}
    \caption{(a) Outdoor vineyard with occluded rigid branches and  trellis that make automation challenging (b) SonicBoom, attached as the robot end-effector link, enables the robot to localize contact using acoustics (\textcolor{red}{red}) as the arm swings into collision in an occluded area (c) Using only acoustics to localize contact points (\textcolor{green}{green}), the robot can interactively map out the occluded object. (d) SonicBoom contains six contact microphones (\textcolor{blue}{blue}) enclosed in a PVC pipe to capture vibrotactile signals (e) observed audio signal from the collision. }
    \label{fig:title_pic}
    \vspace{-5pt}
\end{figure}
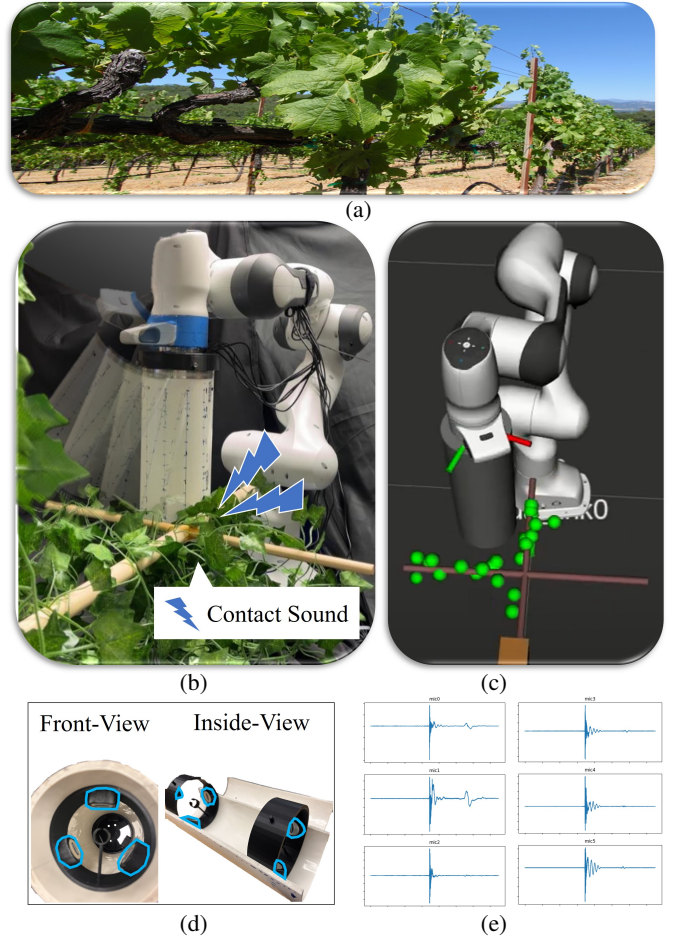

\IEEEPARstart{H}{umans} rely on the sense of touch to navigate and understand cluttered environments where visual perception is limited or unreliable. For instance, farmers picking fruits or pruning branches often use their hands to \emph{feel through the softer canopies} to locate rigid branches by touch. In contrast, robots struggle to incorporate tactile sensing effectively in such unstructured and contact-rich tasks, despite the recent advancements in tactile sensing.
This limitation poses challenges in automating tasks like vine pruning or apple picking, where visual occlusions make detecting branches and trellis difficult (Fig. \ref{fig:title_pic} (a)). Collision with these occluded rigid objects can be hazardous unless carefully incorporated into the robot's motion planner. 
How can robots \emph{feel through occlusion} to safely locate rigid objects when operating in unstructured and harsh outdoor conditions? 

Tactile sensing has shown promise in enabling robots to function in cluttered and uncertain environments~\cite{lin2021exploratory}. However, deploying conventional tactile sensors such as e-skin or gel-based camera sensors over large areas of robotic manipulators remains impractical~\cite{mandil2023tactile}. These sensors are often complex to fabricate, calibrate, and maintain due to wear-and-tear~\cite{sipos2024gelslim}, especially in contact-rich applications like agriculture. For agricultural tasks, where a robotic end-effector must navigate cluttered spaces and endure frequent contact, existing solutions like robotic skins or camera-based tactile sensors may not be sufficiently durable~\cite{lambeta2020digit} while only providing limited coverage~\cite{yuan2017gelsight}. A compelling alternative is to leverage acoustic signals for tactile sensing~\cite{lu2023active}. Vibrations propagate through materials, allowing microphones to capture signals from a large region using a few strategically placed contact microphones~\cite{gandhi2020swoosh}. Furthermore, microphones can be easily protected by embeddeding them within enclosures, such as robot links, that protect the microphones from external wear while enabling practical sensing for contact-rich tasks~\cite{zoller2020active}.

Despite their advantages, prior works on using contact microphones for robotic manipulation have focused primarily on classification tasks, such as classifying coarse contact regions~\cite{brock2022_braile}, material properties~\cite{brock2023_journal}, or number of objects in an enclosed box~\cite{liu2024sonicsense}.
While effective for these classification tasks, prior usage of acoustics in robot manipulation rarely addresses the problem of contact point estimation. In contrast to single-microphone setups that cannot distinguish sounds from different directions or depths, array of multiple microphones enable triangulation by leveraging relative information between sensors, similar to traditional sound source localization (SSL) \cite{desai2022review}. 

However, while SSL methods typically rely on analytical models assuming uniform propagation media like air or elastomers, 
contact-based localization through robot structures presents unique challenges. Vibrations through non-uniform structures like a robot end-effector-link exhibit complex behavior as signal propagates. Propagation paths are affected by geometric irregularities, structural discontinuities, and material changes that cause wave scattering and mode conversions \cite{park2022biomimetic}. 

To address these challenges, we propose SonicBoom, a holistic design of both hardware and data-driven framework that uses an array of contact microphones distributed across the robot end-effector link to localize contact. Using audio features from six microphones, we can precisely estimate contact locations. To provide insight into what features are useful for the tasks of contact localization, we present a detailed analysis of determining the appropriate audio representation and pre-processing methods. Finally, we demonstrate our method on a Franka robot in a mock tree-canopy setting inspired by agricultural tasks (Fig. \ref{fig:title_pic}b).

In summary, our contributions are:
\begin{itemize}
    \item We present SonicBoom, a framework that transforms a robot's end-effector-link into a contact-aware surface by combining an array of microphones with a learning-based approach for contact localization. 
    \item Extensive analysis revealing which audio features and input modalities (spectrogram, phase, robot proprioception) improve the model's capability to generalize.
    \item Robot demonstrations in two real-world settings: robot-active haptic mapping of a mock tree-canopy with 2.0 cm error, and robot-stationary contact localization when human strikes the end-effector with 2.2 cm error.
\end{itemize}

\section{Related Works}~\label{sec:related}
\vspace{-20pt}

\subsection{Robot Acoustic Sensing}
Acoustic sensing has garnered increasing attention recently for robotic applications because microphones are affordable and easily scalable~\cite{yoo2024poe,park2022biomimetic}. By using one or more microphones, researchers have demonstrated the ability to perceive dynamic physical scenes such as balls bouncing on tables~\cite{matl2021stressd}, estimate soft body deformations~\cite{yoo2024poe}, observe movements of objects in a box~\cite{gandhi2020swoosh, chen2021boombox} and contact estimation~\cite{brock2022_braile}.

As shown by these works, sound data are information-rich despite their simplicity. To extract useful features from time series signals from the microphones in these contexts, researchers have proposed various feature engineering techniques. Commonly used features are time shifts among the microphones~\cite{matl2021stressd}, fast Fourier transform (FFT) features~\cite{perrodin2012design}, and spectrograms~\cite{gandhi2020swoosh, cabrera2019detection}. Based on the choice of signal processing methods, the extracted features may filter out noise or useful information. For example, the window size for extracting spectrograms needs to be tuned to balance the trade-off between temporal and frequency resolution~\cite{nisar2016efficient}. 

While existing work has explored task-based representations ~\cite{gandhi2020swoosh} and self-supervised embeddings ~\cite{thankaraj2023sounds} for acoustic signals, the impact of specific signal processing steps and feature representations on downstream robotic perception tasks remains less explored. In this work, we contribute an in-depth analysis of the various feature representations of the acoustic signals for the task of robot manipulator contact estimation. 


\subsection{Acoustic Sensing for Contact Estimation}
A compelling application for robotic acoustic sensing is contact estimation~\cite{park2022biomimetic}. Microphones are relatively easy to install on robotic systems~\cite{yoo2024poe} and can observe contact by using the robot as the medium for vibration caused by contact~\cite{park2022biomimetic}.

Researchers have approached contact estimation on robotic manipulators with both active and passive acoustic sensing methods. In active acoustic sensing methods, a sound source is embedded with the microphones~\cite{zoller2020active}. By observing the change in the known sound source signals, researchers have demonstrated the ability to sense contact and even sense objects' proximity~\cite{rupavatharam2023ambisense}. However, these works often require proprietary hardware for an active sound source to sweep frequencies ~\cite{SamsungAI_2023sonicfinger, SamsungAI_fan2022}. These methods are also not yet scalable because sensorizing large surfaces would require multiple sound sources due to signal attenuation, and no work has demonstrated a method to eliminate cross-talk problems in multi-sound-source active acoustic contact estimation systems.

Passive acoustic tactile sensing benefits from its system simplicity since only microphones need to be embedded~\cite{park2022biomimetic} and it avoids sensor cross-talk challenges. However previous works have only demonstrated spatially coarse contact estimation with one microphone~\cite{zoller2018acoustic} or contact estimation on well-controlled flat surfaces where analytical methods such as triangulation are sufficient without the added challenges of handling robot noises or complex vibration propagation behaviors on non-flat surfaces~\cite{park2022biomimetic}. To the best of our knowledge, SonicBoom presents the first work on using an array of microphones on a robotic end-effector to perform state-of-the-art contact localization that demonstrates generalization to various settings out-of-distribution from the training set.




    

 %

\section{SonicBoom Hardware}~\label{sec:overview}

The SonicBoom end-effector design consists of a 4" x 12" (radius x height) PVC pipe housing six piezoelectric contact microphones arranged in two rings of three sensors each, positioned at both ends of the tube (Fig. \ref{fig:title_pic}d). We strategically chose PVC for its lightweight properties and higher damping coefficient compared to aluminum, which helps create more distinct signals across microphones based on proximity by absorbing vibration energy during propagation. The contact surface can be parameterized in a 2D space defined by height $z$ and azimuth angle $\theta$, where each three-microphone ring enables spatial localization through triangulation\cite{sensors_survey_2023}. The two-ring configuration provides overlapping coverage regions for redundant sensing while extending the contact-aware surface along the entire length of the end-effector. This simple design enables easy extension of the sensing region by adding more rings according to the pipe length, making it adaptable for longer end-effector links such as when reaching for apples on taller trees.
 
To facilitate assembly and sensor mounting, the tube is split longitudinally with pre-drilled mounting holes. Although this structural modification introduces additional non-uniformity in signal propagation, it is necessary to mount sensors. The system uses passive piezoelectric microphones that convert mechanical vibrations into electrical signals, which are then amplified and digitized through an eight-channel Behringer UMC1820 DAQ at 44.1 kHz sampling rate. The end of the tube is equipped with an interchangeable tool mount that can accommodate, e.g., a pneumatic suction cup or a single-actuated pruner.

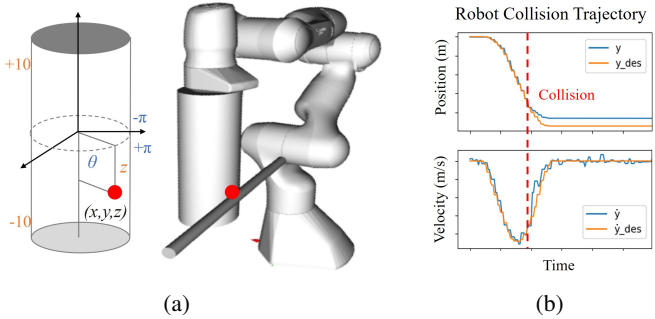
\begin{figure}[t]
    \centering
    \includegraphics[width=1\linewidth]{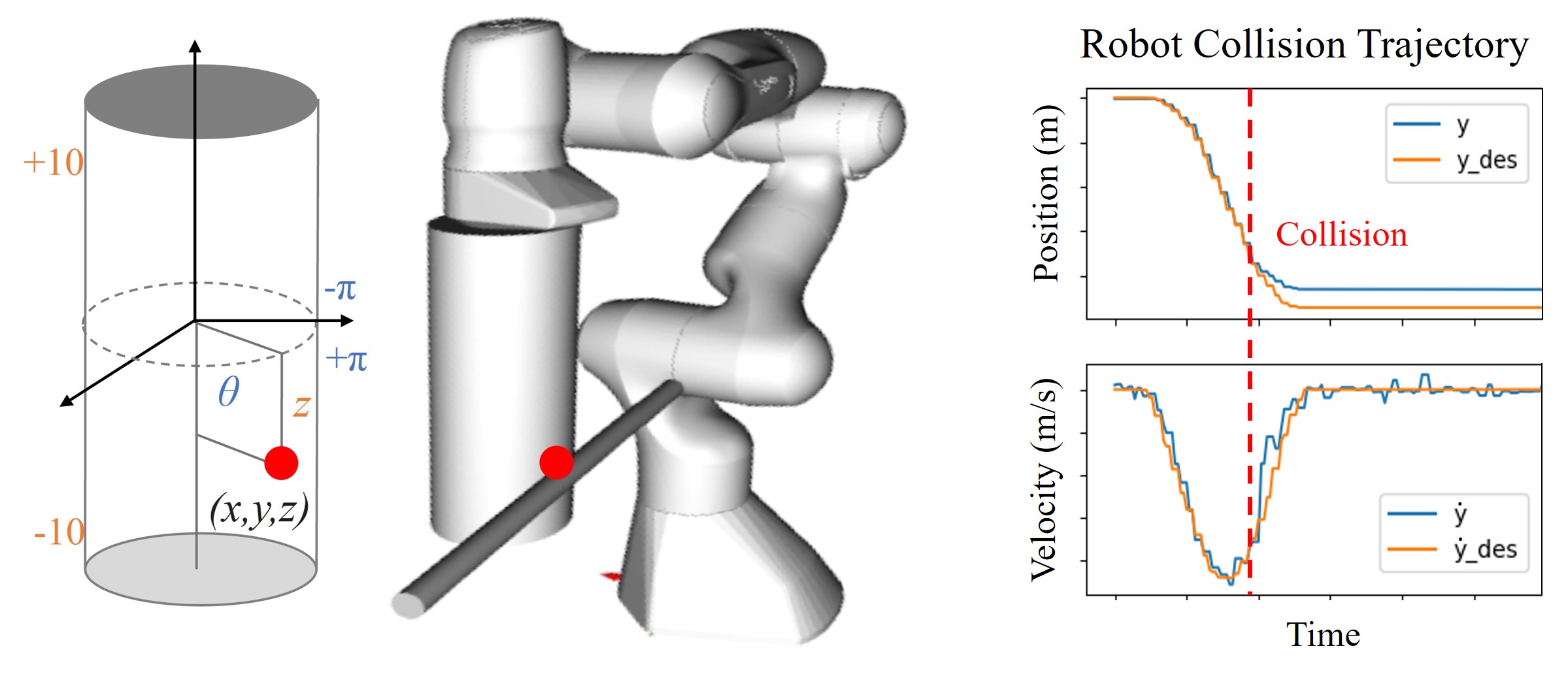}
    \vspace{-5pt}

    \setlength{\unitlength}{1cm}
    \begin{picture}(0,0)
    \put(-2.2,0.3){\textcolor{black}{\small (a)}}
    \put(2.75, 0.3){\textcolor{black}{\small (b)}}

    \end{picture}

    \vspace{-5pt}
    \caption{ (a) SonicBoom end-effector link parametrized in cylindrical coordinates (b) Striking motion used to create collision acoustic signals are shown with end-effector position and velocity profile.}
    \label{fig:hw_trajectory}
    \vspace{-5pt}
\end{figure}

\section{Problem Formulation}~\label{sec:problem}

When SonicBoom end-effector collides with a rigid object, the impact generates vibrations that propagate through the material and are captured as acoustic signals $A$ by the microphone array. The core problem is to predict the contact location $p$ on SonicBoom's end-effector surface using these acoustic signals $A$ and the robot's motion trajectory $X$ as inputs. We parameterize the contact location of SonicBoom in cylindrical coordinates $p(z,\theta)$ using azimuth angle $\theta \in [-\pi, +\pi]$ and height $z \in [-10, +10]$ cm, with the cylinder's mid-point of the central axis as the reference (Fig. \ref{fig:hw_trajectory}a). This formulation can be expressed formally as learning a mapping function $f(A, X) \rightarrow p(z,\theta)$. 

For tractability, we simplify the problem by focusing on contact events with two key constraints: impulsive contacts (e.g., strikes) rather than continuous contacts (e.g., sliding), and single-point contacts rather than simultaneous multi-point contacts. These constraints align well with our target agricultural applications, where continuous sliding along a rigid branch is both mechanically risky and damaging to the vegetation. Furthermore, tree branches can be geometrically approximated as cylinders \cite{kim2024_treegnn}, where cylinder-to-cylinder collisions naturally result in single-point contacts, making these assumptions particularly suitable for our problem domain.

\section{Dataset Generation}~\label{sec:dataset}
\vspace{-10pt}

Unlike common datasets for speech recognition or sound classification, contact-based acoustic data is scarce and challenging to simulate accurately due to the physics behind energy propagation through non-uniform medium and geometric irregularities \cite{park2022biomimetic}. Moreover, contact interactions exhibit complex dependencies between contact properties and their acoustic signals - even striking the same point on SonicBoom's end-effector with a wooden rod produces varying vibration patterns depending on the rod's contact location, as it behaves like a cantilever beam with distinct oscillation modes. To address these challenges, we develop an automated data collection pipeline using a Franka robot equipped with our SonicBoom end-effector, systematically capturing acoustic signatures from various striking actions and beam objects in real-world collision data. Our dataset pairs six-channel audio signals and robot proprioceptive data with contact locations on the SonicBoom surface $p(z,\theta)$.

\subsection{Data Collection Process}
Our dataset comprises over 18,000 collision events collected through an automated process where the robot strikes rigidly mounted wooden rods with the SonicBoom end-effector. The cylindrical geometries of both the rod and end-effector ensure single-point contacts. Striking motions are generated by setting a desired goal pose past the object using an position controller with low impedance gain at the contact joints and high gain at other joints. For each interaction, we record a two-second audio window capturing the complete sequence: approach trajectory, collision, and subsequent damping response (Fig.\ref{fig:hw_trajectory}).

Given that trees can be geometrically approximated as cylinders\cite{kim2024_treegnn}, we use wooden cylindrical rods of varying lengths and thicknesses to create our dataset. Our training set includes four wooden rods with different dimensions, while the evaluation set uses six rods that are held-out. We further increase dataset diversity by varying both strike velocity and strike angle. The complete dataset contains 108,000 audio files from 18,000 contact events, collected over approximately 100 robot hours.

\subsection{Obtaining Labels for Contact Point}\label{sec:gt}

We obtain ground-truth contact locations through post-processing of the recorded robot trajectories. The process involves loading robot joint states at collision time into a robot mesh representation along with the 3D pointcloud of the contact object obtained using FARO's 3D laser scanner. At collision, naively choosing the closest point between the robot and object leads to incorrect contact point identification due to mesh intersection issues. Therefore, if there are numerous near-zero intersection points, we average all the points and then project the averaged point on to the SonicBoom end-effector surface to handle potential errors in the label.

\section{Audio Localization}~\label{sec:method}
\vspace{-10pt}

\begin{figure}[t]
    \centering
    \includegraphics[width=1\linewidth]{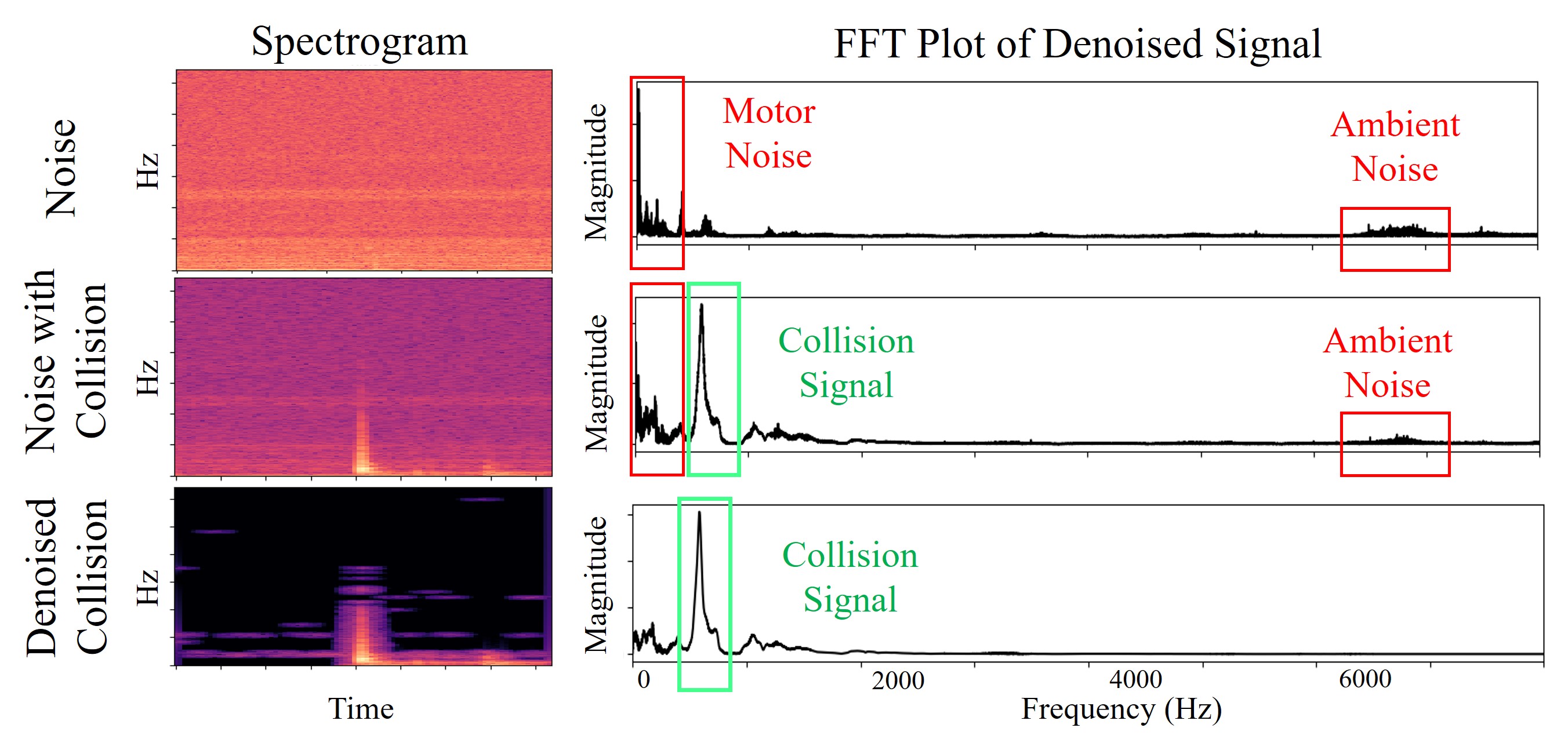}
    \vspace{-15pt}
    \caption{Frequency analysis to de-noise the collision signal. 
    The motor noise and ambient noise can be isolated to specific frequency regions, and filtered to obtain a clean collision signal. }
    \label{fig:method_signal}
\end{figure}

\begin{figure*}[t]
    \centering
    \includegraphics[width=1\linewidth]{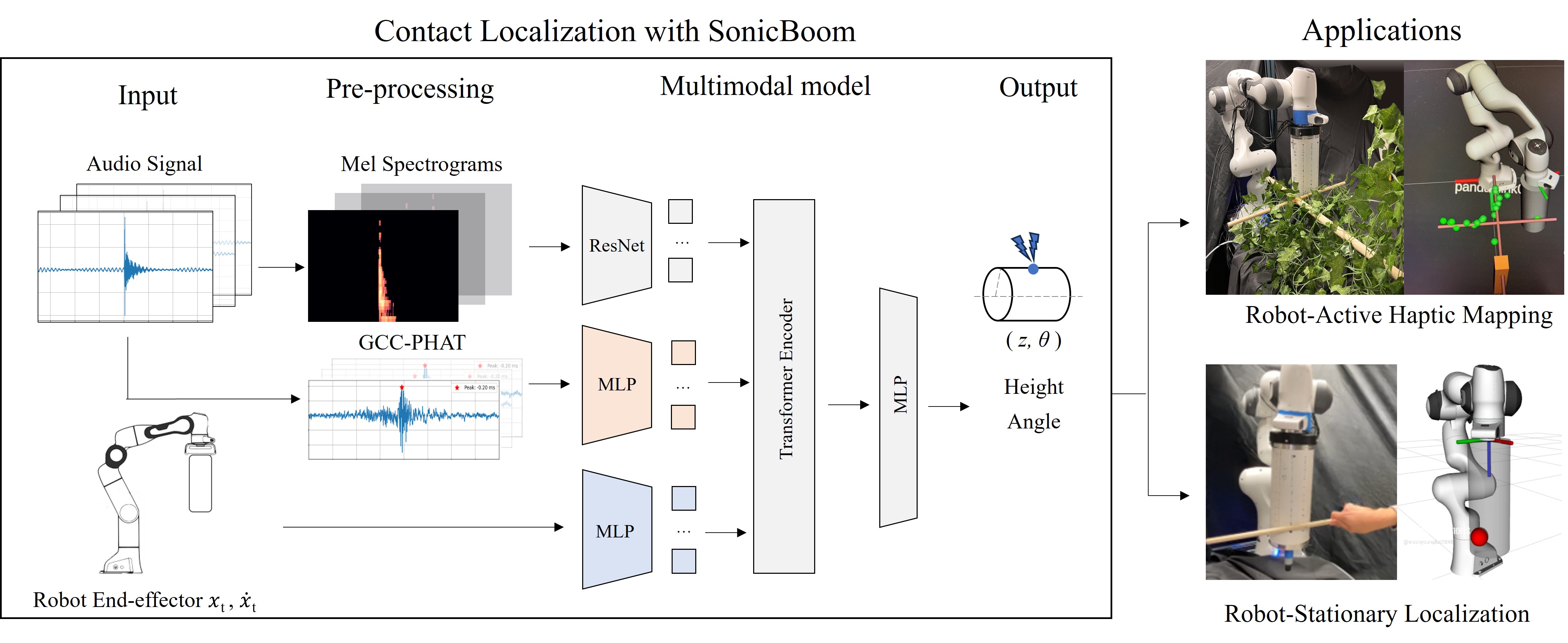}
    \vspace{-15pt}
    \caption{System overview of SonicBoom for contact localization in two settings. The inputs used for localization 
    are audio and robot proprioceptive data. Audio signal is pre-processed into mel spectrograms and GCC-PHAT. Each sensing modality is encoded into a latent feature before being fused by the multi-sensory self-attention transformer encoder. The output prediction is represented in cylindrical coordinate $z,\theta$ along SonicBoom surface, which can be used for haptic mapping or localization.}
    \label{fig:sw_overview}
\end{figure*}

\subsection{Preprocessing Audio Signals}\label{Preprocessing}
We process the raw audio signals through several stages before training to ensure data quality and consistency. First, all six microphone channels are recorded into a single Waveform Audio file (WAV) to ensure temporal alignment. We then implement spectral gating to filter out background noise using a pre-recorded reference signal of robot motion without collisions, effectively isolating the collision signal from both motor and ambient noise (Fig. \ref{fig:method_signal}). The denoised signals are trimmed to one-second windows centered around the peak amplitude to capture the relevant collision event.

We then convert the preprocessed time-domain signals to mel spectrograms using Short-Time Fourier Transform (STFT) with an FFT window size of 512, hop length of 128, and 50 mel filter banks. 
Lastly, we normalize each microphone channel's spectrogram independently for each channel across trials, frequency bins, and time steps. By normalizing each channel individually rather than jointly across all channels, we preserve the relative signal differences between microphones that are crucial for localization.

\subsection{Input Features}
Traditional Sound Source Localization (SSL) methods typically operate in environments where sound propagates through homogeneous media like air or liquid, with sources located significantly farther than inter-microphone distances. However, contact-based localization presents challenges due to faster vibration propagation through solid media, complex geometric structures affecting propagation, and violation of far-field assumptions common in traditional SSL. 
While one audio feature may be susceptive to noise or lack resolution, our hypothesis is that fusing multi-modal features that complement each other can improve robustness in spatial mapping. We identify three key representations that provide complementary information for contact localization.\\

\textbf{Mel Spectrogram}:
Mel spectrograms capture the energy distribution across frequency and time by computing the power spectrum of overlapping windows of the signal and mapping them to mel-scale frequency bins. By discarding phase information and preserving magnitude, mel spectrograms effectively represent the intensity patterns of contact events and their resonance responses. However, this high-dimensional representation is for each individual microphone, and do not explicitly encode the relative timing differences between signals that are crucial for localization.
\\

\textbf{GCC-PHAT}:
Generalized Cross-Correlation with Phase Transform (GCC-PHAT) \cite{sensors_survey_2023} explicitly computes similarity between microphone pairs as a function of time-lag. This representation improves robustness against noise and reverberation by normalizing the cross-power spectrum to have unit magnitude at all frequencies, GCC-PHAT emphasizes phase alignment while being robust to amplitude variations between microphones \cite{he2018GCC}. This weighting is particularly valuable for low-cost uncalibrated piezoelectric sensors, as it eliminates the need for calibrating the gains across different microphones. We compute the GCC-PHAT for all 15 possible pairing between the six microphone. Given two time-domain audio signals \(x_i(t)\) and \(x_j(t)\), the GCC-PHAT is defined as:

\begin{equation}
    \hat{G}_{\text{PHAT}}(f) = \frac{X_i(f) X_j(f)^*}{|X_i(f) X_j(f)^*|}
\end{equation}
Where \(X_i(f)\) and \(X_j(f)\) are the Fourier transforms of the two signals and \([\cdot]^*\) denotes the complex conjugate. The Time Difference of Arrival $\hat{d}$ for these two microphones is estimated by finding the peak value as:

\begin{equation}
    \hat{d}_{\text{PHAT}}(i, j) = \underset{d}{\arg\max} \left( \hat{R}_{\text{PHAT}}(d) \right)
\end{equation}
Where \(\hat{R}_{\text{PHAT}}(d)\) is the real part of the inverse Fourier transform of $\hat{G}_{\text{PHAT}}(f)$. Empirically, we found that naively using the extracted $\hat{d}$ resulted in poorer performance due to noise in the signal. Therefore we opt to use the less explicit and full representation of GCC-PHAT vector as similarly done in \cite{he2018GCC}. \\

\textbf{Robot Proprioception}:
Direction of motion provides a strong prior for contact localization. The intuition is straightforward yet effective: collision is likely to occur in the direction of robot's motion and highly unlikely on the opposite side, particularly when interacting with static and inanimate objects. We use one second trajectory of the end-effector's pose and velocity time is time-aligned with the audio collision signal. This contextual information helps constrain the possible contact locations based on the robot's trajectory.

\subsection{Learning Contact Localization}
Our dataset $\mathcal{D}$ consists of audio and proprioceptive samples paired with contact location samples, represented as
$\mathcal{D} \equiv$ $\left\{A_n, G_n, X_n,  p_n\right\}_1^N$ where ${A_n} = [A_1,...,A_6]$ represents six acoustic signals as mel spectrograms, ${G_n} = [G_1,...,G_{15}]$ contains the GCC-PHAT features between microphone pairs, ${X_n}$ captures the robot end-effector trajectory (pose and velocity), and  ${p_n} = [z_n, \theta_n]$ denotes the contact point on the SonicBoom surface. 

Through supervised learning, our model learns to predict contact points as $[\hat{z}_n, \hat{\theta}_n] = f(A_n, G_n, X_n)$. We stacked the pre-processed mel spectrogram $A_n$ from all six microphones to create a multi-channel representation,  where each channel captures the time-frequency structure of signals from different spatial locations. To improve robustness, we apply data augmentation through time and frequency masking and translation randomization.

For fusing the two audio representations and proprioceptive features together, we adopt a multi-modal transformer architecture similar to \cite{li2022see, liu2024maniwav}, where each sensing modality first passes through a dedicated encoder before fusion in the transformer (Fig. \ref{fig:sw_overview}). For the spectrogram encoder, we empirically found that ResNet50 outperforms Audio Spectrogram Transformer (AST) \cite{gong2021ast} for our setting. While CNN-architectures like ResNet may not capture long-range temporal dependencies like AST, it excels at extracting local patterns crucial for our impulse-like collision signals. Both GCC-PHAT and proprioceptive features are encoded using three-layer MLPs. To emphasize learning from audio signals, we allocate twice the embedding dimension for audio features than proprioceptive features and apply lower dropout rates during training. The embeddings of each encoder are concatenated and passed into the transformer that fuses the modal-specific encodings. The overarching architecture design choices focus on preserving modality-specific characteristics while enabling effective cross-modal learning through the transformer architecture.

For training, we minimize the MSE loss between predicted and ground truth contact points. To handle the circular nature of the azimuth angle $\theta$, we decompose it into Cartesian coordinates $(x, y)$ before computing the loss, avoiding discontinuities at boundary wrapping. 

\begin{equation}
\mathcal{L}_{\mathrm{p}}=\mathbb{E}\left[\left\|f(A,G,X)-p\right\|_2^2\right]
\end{equation}

\begin{equation}
\mathcal{L}_{\mathrm{p}} = \mathbb{E}\left[\left\|
\begin{bmatrix} 
\sin\hat{\theta} \\ 
\cos\hat{\theta} \\ 
\hat{z} 
\end{bmatrix}
-
\begin{bmatrix} 
\sin\theta \\ 
\cos\theta \\ 
z 
\end{bmatrix}
\right\|_2^2\right]
\end{equation}


We train the model for 200 epochs using the Adam optimizer with a batch size of 64 and a learning rate scheduler (initial LR = 0.001) on an NVIDIA RTX 3090 GPU, requiring approximately four hours of training time.


\section{Experimental Results}~\label{sec:experiments} 
\vspace{-10pt}

To investigate whether our method of contact localization can generalize, we designed our experiments to have the robot collide into novel objects as well as with novel striking actions not seen during training. We quantify error  using Mean Euclidean Distance (MED) between the predicted contact point and the ground truth point obtained from mesh collision in 3D space (discussed in \ref{sec:gt}). Our aim to investigate our system through three complementary analyses that address the questions:
(1) \textit{Can SonicBoom generalize to novel acoustic signals from out of distribution contact events?} (2) \textit{From the multi-modal inputs, what features are essential for robust localization?}

\subsection{Evaluation on Localization}\label{sec:eval_localization}

 \begin{figure}[t]
    \centering
    \includegraphics[width=1\linewidth]{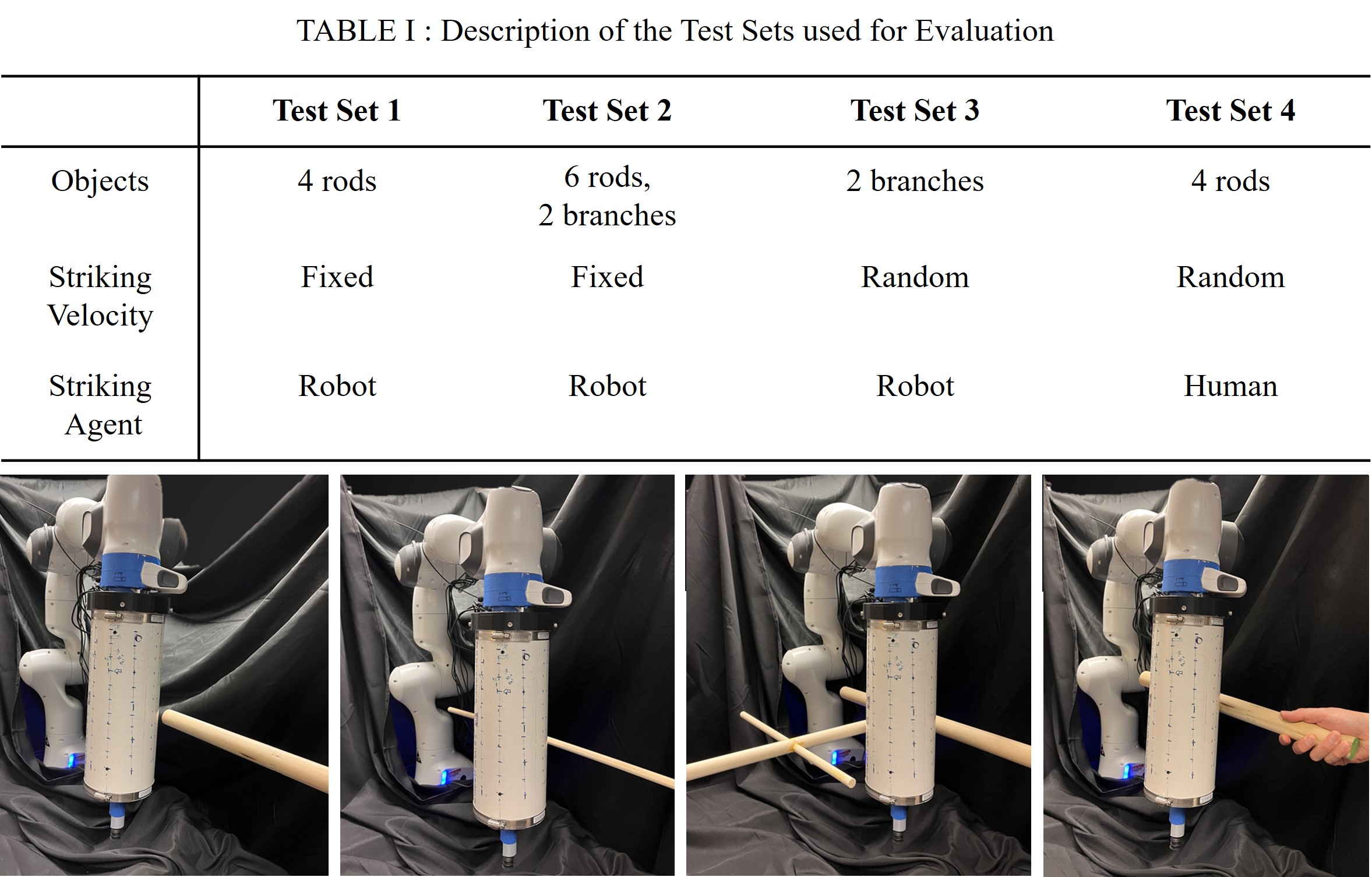}
    \vspace{-15pt}
    
    \caption{Evaluation sets increase in complexity as objects, striking velocity, and striking agents are varied. Training set is composed of variations of simpler single-rod sticks while evaluation set is composed of novel wooden rods and complex tree-like geometric structures.}
    \label{fig:eval_set}
    \vspace{-5pt}
\end{figure}

\begin{figure}[t]
    \centering
    \includegraphics[width=1\linewidth]{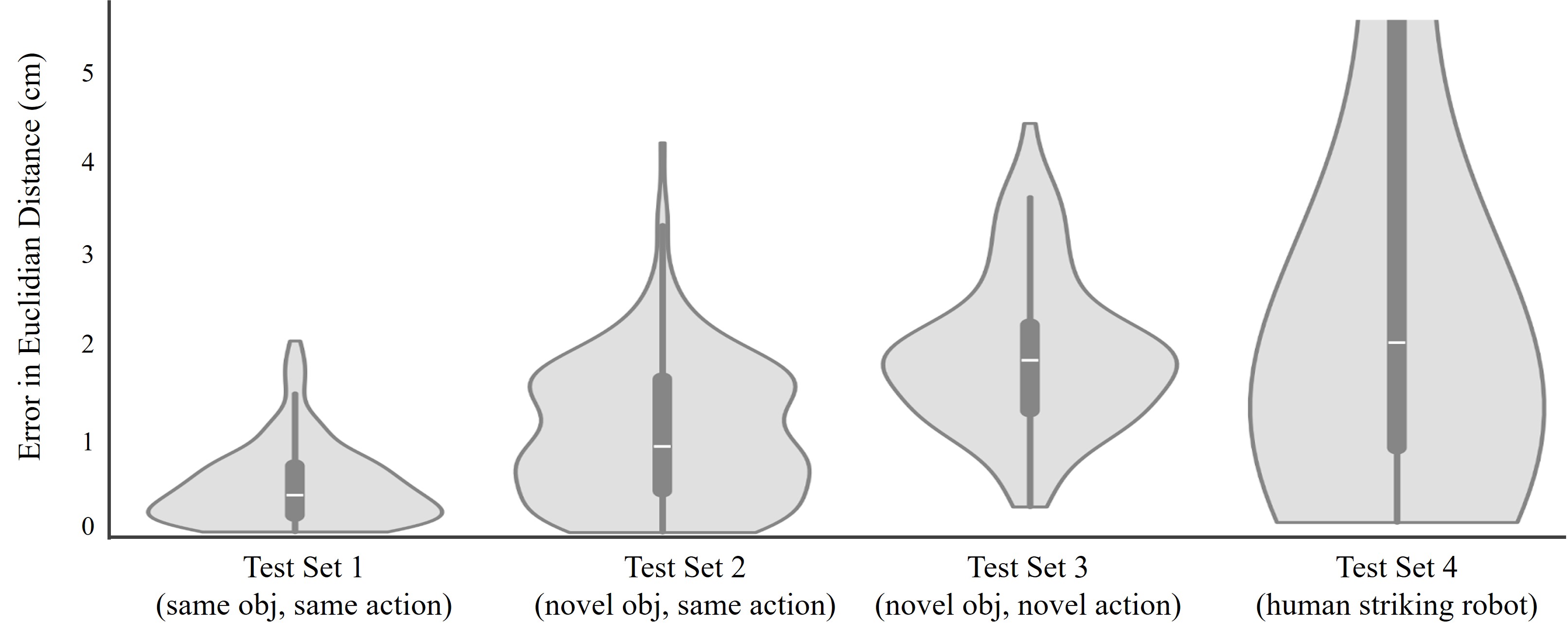}
    \vspace{-15pt}
    \caption{Error distribution in localization visualized with box-whisker plots combined with violin plots. Error increases with increasing complexity in test sets, while maintaining practical performance. Y-axis is clipped at 6cm for improved readability, full range can be seen in Fig. \ref{fig:input_ablation} (\textcolor{myYellow}{yellow}). }
    \label{fig:localization_plot}
    \vspace{-5pt}
\end{figure}

We evaluate SonicBoom's contact localization capabilities through four progressively challenging test scenarios (Fig. \ref{fig:eval_set}) with over 200 samples each. Across the test sets, the error distribution  (Fig. \ref{fig:localization_plot}) shows increasing errors of 0.43, 1.01, 2.01, and 2.22 cm, reflecting the increasing complexity while demonstrating generalizing beyond its training distribution to handle novel objects and contact scenarios.

\textbf{Test Set 1} is our validation set where we use identical wooden rods and tapping motions as in the training data. The model prediction yields a MED of 0.43cm, validating the model's ability to accurately map vibrotactile signals to contact locations under controlled conditions. This implies that the input audio data contain \textit{rich} information for localization, and that the utilized model architecture is sufficiently expressive for our task. 

\textbf{Test Set 2} is to evaluate novel wooden geometries, including rods of different dimensions and tree-like structures (Fig. \ref{fig:eval_set}). Despite training on only four wooden rods, the model maintains strong performance with a MED of 1.01 cm across six wooden rods with unseen dimensions and to novel geometric variations similar to mock branches, suggesting that SonicBoom can generalize to different geometries as long as they share similar material properties of the wooden rod. 

\textbf{Test Set 3} contains collision samples collected during robot-active haptic mapping of the mock tree canopy. This scenario is most similar to how SonicBoom would be used for real-world deployment. It is also the most challenging set because it introduces novel contact motions used for random exploratory movements, as well as noise for the audio from bristling of leaves. This exploration strategy covers a significantly larger robot workspace than the training set, producing striking velocities and contact angles that vary from the predetermined tapping motions used in training. Despite these challenging conditions, SonicBoom maintains robust performance with a MED of 2.01cm. This section is discuss further in Sec. \ref{sec:mapping}.

\textbf{Test Set 4} presents our most challenging evaluation scenario by isolating the contribution of acoustic sensing from robot proprioception. While our model was trained on paired audio and proprioceptive data, we now test its performance when proprioceptive information is unavailable by keeping the robot stationary and having a human strike the end-effector with wooden rods of varying thicknesses. This zero-shot transfer scenario forces the model to rely solely on acoustic signals for localization. Despite never encountering such interaction patterns during training, SonicBoom maintains robust performance with a MED error of 2.22 cm. We analyze this result further in Sec. \ref{sec:drum}.


\begin{figure}[t]
    \centering
    \includegraphics[width=1\linewidth]{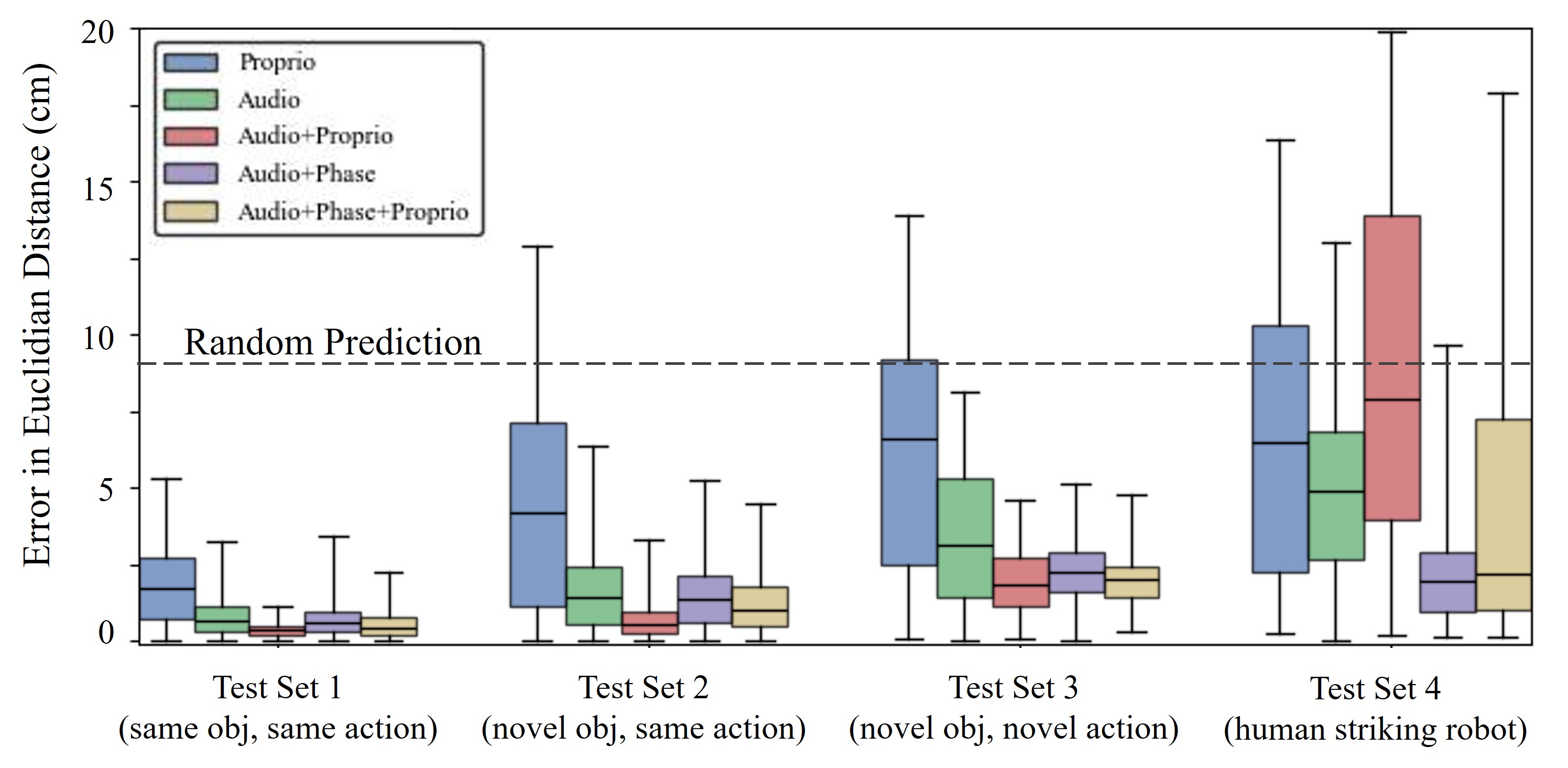}
    \vspace{-5pt}
    \caption{Ablation study of varying input modalities across varying difficulty of test sets. Audio+phase (\textcolor{myPurple}{purple}) outperforms other combinations of multi-modal features in generalization, where it can even zero-shot transfer to human striking the SonicBoom.}
    \label{fig:input_ablation}
    \vspace{-5pt}
\end{figure}

\subsection{Ablation Study}\label{sec:Augmentation}
 We conduct ablation studies examining input modalities, signal representations, and preprocessing methods to reveal insights about which features implicitly captures spatial-temporal information helpful for localization.

\subsubsection{Input Modality}

We evaluate different combinations of input modalities on localization performance across our four test sets. Fig. \ref{fig:input_ablation} presents the error distributions using box-whisker plots (z-score = 2), where boxes indicate quartiles and whiskers represent the minimum and maximum values within the statistically significant range.

Generally, models using only single modality  (\textcolor{blue}{blue}, \textcolor{green}{green}) performed the worst across the test sets. 
Combining audio and proprioception (\textcolor{red}{red}) resulted in the lowest error in test sets 1,2 but increased in error for more complex sets, even having the highest error in test set 4, which is indicative of overfitting to proprioception data during training. 
Importantly, models utilizing phase input (\textcolor{myYellow}{yellow}, \textcolor{myPurple}{purple}) maintained low error even on the most out of distribution test set 4, implying importance of phase for generalization. It is important to note that test set 4 is purposefully designed to investigate whether the models are actually learning from audio or heavily relying on proprioception. As test set 4 doesn't contain useful proprioception data, models only using audio (\textcolor{green}{green}, \textcolor{myPurple}{purple}) maintained relatively similar performance compared to test set 3, while models using proprioception (\textcolor{red}{red}, \textcolor{myYellow}{yellow}) observed significantly higher errors.
Between the two models that use phase, the combination of all three modalities (\textcolor{myYellow}{yellow}) achieved better performance on test set 1,2,3. This suggests each modality contributes complementary information for contact localization.

Examining the single-modal using proprioceptive (\textcolor{blue}{blue}) produced an insightful result as well. As expected, direction of motion can be a strong prior to which edge of the end-effector would come in contact (i.e. swinging a human arm outwards, expect to make contact on outside edge of arm). Although error on proprioceptive is high across all test sets in Fig. \ref{fig:input_ablation}, further decomposing Euclidean distance into height and angle showed proprioceptive data was greatly effective in predicting contact angle (11.7° error) but poor in estimating height (4.3 cm error). This aligns with our intuition that proprioception provide strong directional cues but limited information about contact height along the cylindrical surface.

\subsubsection{Preprocessing and Augmentation} 
\begin{figure}
    \centering
    \includegraphics[width=0.9\linewidth]{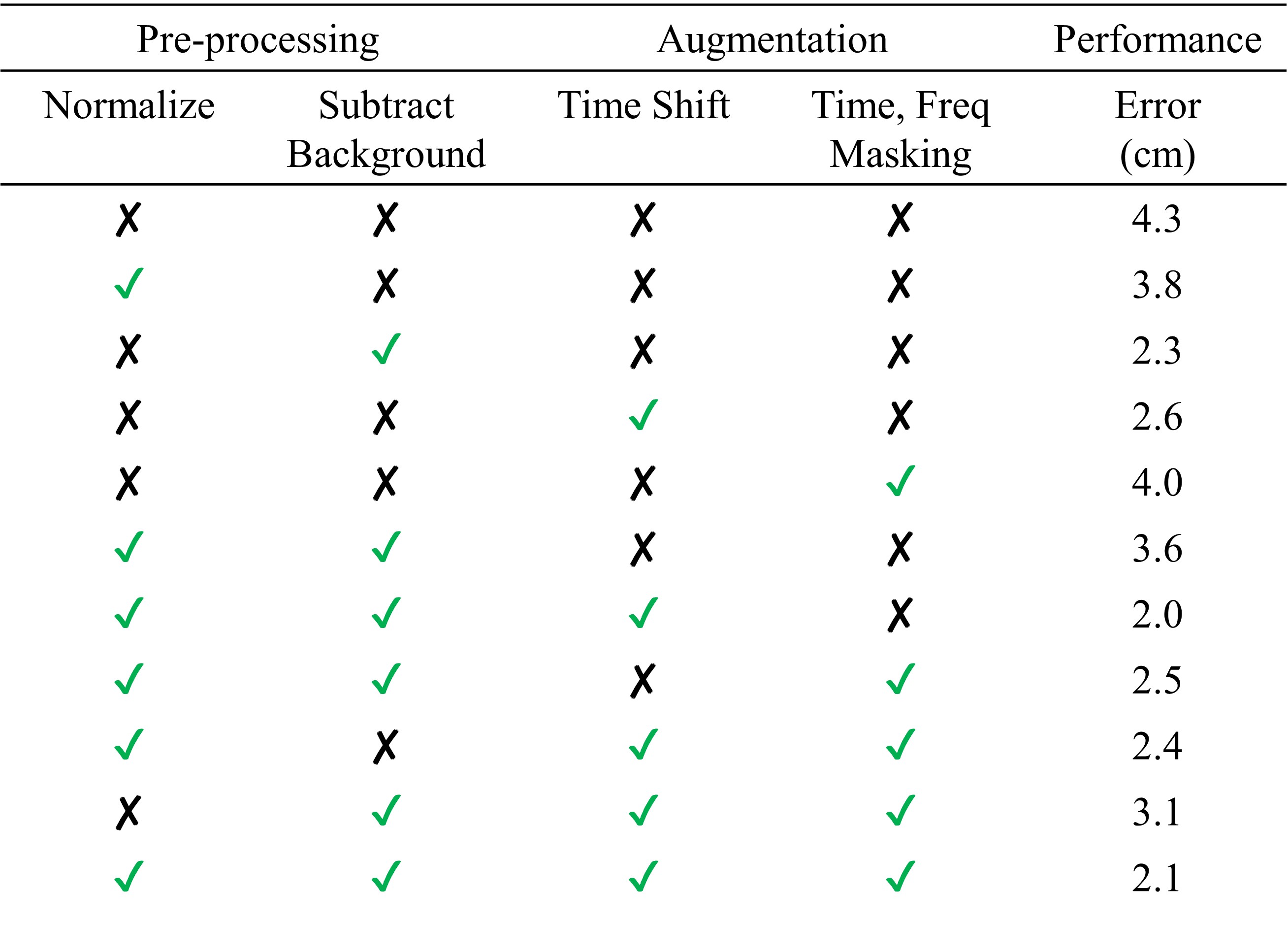}
    \vspace{-5pt}
    \caption{Ablation  of preprocessing and augmentation methods for audio.}
    \label{fig:Ablation_table_localization}
    \vspace{-5pt}
\end{figure}

Using the model that only uses audio input, we conduct an additional ablation study. The pre-processing and augmentation methods used to train individual models are shown in Fig. \ref{fig:Ablation_table_localization}. To facilitate rapid experimentation, we employ a lightweight CNN architecture and evaluate different combinations of preprocessing and augmentation techniques on test set 1. Our augmentation utilizes temporal translations while preserving frequency information, as shifting contact timing helps generalization while frequency shifts could degrade localization performance. Results show that background subtraction and time shifting steps improve localization accuracy most, while time-frequency masking produces minimal impact.

\subsection{Real-World SonicBoom Demonstration}\label{sec:3D_reconstruction}

\subsubsection{Robot-Active Haptic Mapping}\label{sec:mapping}

\begin{figure}
    \centering
    \includegraphics[width=0.9\linewidth]{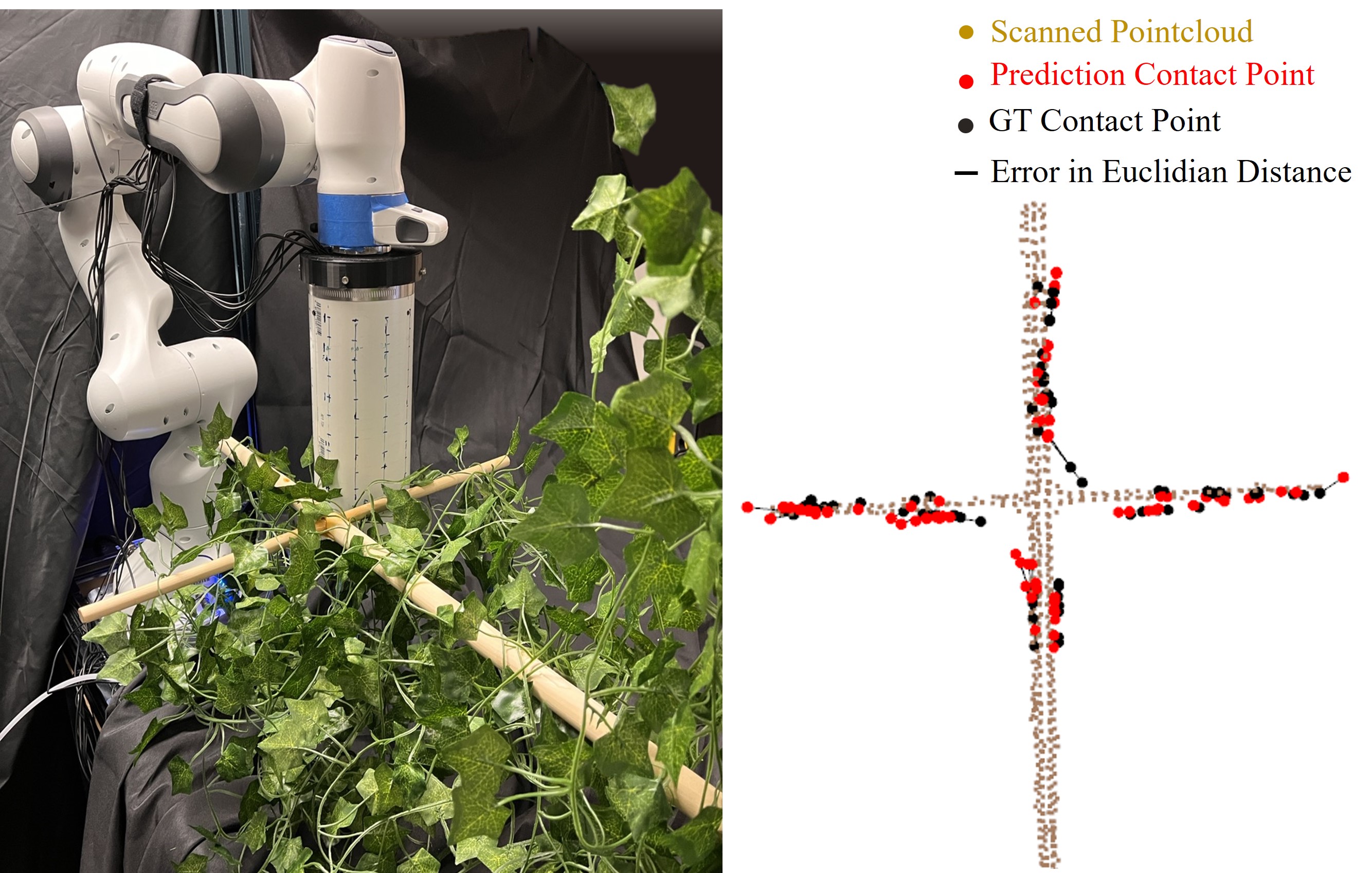}
    \vspace{-5pt}
    \caption{Demonstration of haptic mapping in cluttered environment where leaves visually occlude the rigid branch but SonicBoom allows to localize with contact. }
    \label{fig:mapping}
    \vspace{-5pt}
\end{figure}

We demonstrate SonicBoom's practical utility through haptic mapping in occluded spaces inspired from robot arm reaching through cluttered branches in an agriculture setting (Fig. \ref{fig:mapping}). The system detects wooden branches in 3D space despite heavy occlusion from artificial vines, which introduce additional contact noise through bristling and brushing sounds. 

To generate contact, the robot employs a simple exploration motion used for haptic mapping. We develop a guided sampling strategy based off the object's scanned pointcloud. Rather than having the robot execute completely random motions in 3D space, we sample positions uniformly in a 2D plane around the object, and filter out positions that would result in mesh intersections between SonicBoom and the object. From these valid sample positions, the robot executes striking motions in four directions in $x,y$ plane (left, right, up, down). From these random striking motions, we add only the interaction event that results in an audio signal exceeding a predetermined amplitude threshold.

We evaluate the mapping accuracy using two metrics. First, we compute MED between corresponding points in the predicted pointcloud $\hat{P}$ (\textcolor{red}{red}) and ground truth point cloud $P$ (black), achieving an error of 2.01cm (Fig. \ref{fig:localization_plot}). Second, to compare the two pointclouds, we adopt a unidirectional root mean square variant of the Chamfer Distance:

\begin{equation}
    CD(\hat{P}, P) = \sqrt{\frac{1}{|\hat{P}|} \sum_{x \in \hat{P}} \min_{y \in P} \|x - y\|^2}
\end{equation}

This metric yields a distance of 2.0cm between $\hat{P}$  and $P$.


\subsubsection{Robot-Stationary Contact Localization} \label{sec:drum}

\begin{figure}
    \centering
    \includegraphics[width=1\linewidth]{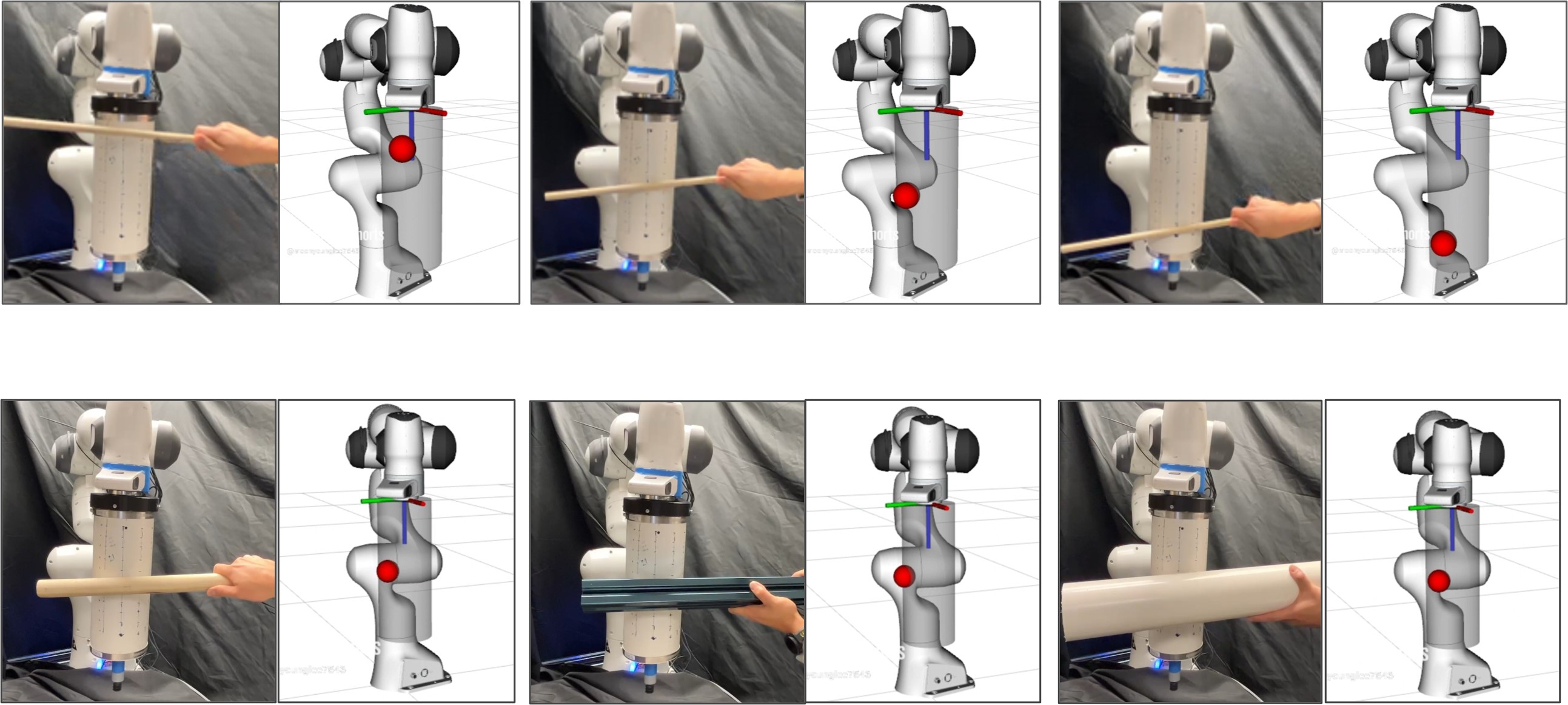}
    \setlength{\unitlength}{1cm}
    \begin{picture}(0,0)
    \put(0, 2.35){\textcolor{black}{\small (a)}}
    \put(0, 0.05){\textcolor{black}{\small (b)}} 
    \end{picture}
    \vspace{-10pt}

    \caption{Zero-shot evaluation for (a) novel contact event where human strikes robot as opposed to robot striking the object, and (b) novel objects with different material properties. Prediction points are shown in (\textcolor{red}{red}).  }
    \label{fig:drum}
    \vspace{-5pt}
\end{figure}
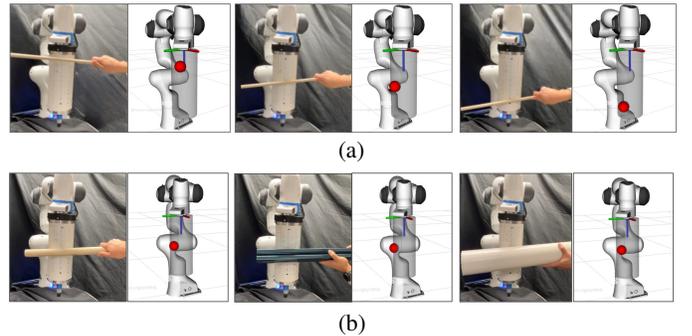

To isolate the contribution of acoustic sensing from robot proprioception, we evaluate the system while stationary as a human strikes various locations on SonicBoom surface. The surface was visually marked with 2cm intervals to guide the strike location, and was struck in sequence, providing ground truth. Using only audio spectrogram and phase information, the model achieves 2.4 cm mean localization error in this zero-shot transfer scenario. Furthermore, we observe generalization to novel materials: testing across 90 strikes with wood, aluminum, and PVC pipes yields mean errors of 2.8, 3.4, and 4.4 cm respectively (Fig. \ref{fig:drum}), demonstrating SonicBoom's potential to transform the end-effector-link to contact-aware surface using microphones.

\section{Conclusion}

We present SonicBoom, a contact localization system using acoustic sensing that enables robot arms to spatially locate collisions with rigid obstacles when navigating cluttered environments. By embedding an array of contact microphones within the robot end-effector-link and developing a data-driven approach to interpret vibrotactile signals, our system achieves precise localization with mean errors of 0.4cm to 2.2cm across increasingly challenging scenarios. Through extensive ablation and real-robot settings, we demonstrated that acoustic-based sensing can provide reliable spatial awareness and surprising generalization ability to novel contact events. While our current work focuses on estimating a single point independent of prior samples, for future work, we will investigate using a probabilistic filtering method to estimate additional contacts given past predictions. Additionally, leveraging the temporal structure of acoustic signals and the sequential modeling capabilities of transformers could enable continuous tracking of sliding contacts, expanding the system's utility for manipulation tasks out in the field.

 %


\section*{Acknowledgment}
This work was supported by NSF Robust Intelligence 1956163 and NSF/USDA NIFA AIIRA AI Research Institute 2021-67021-35329.

\ifCLASSOPTIONcaptionsoff
  \newpage
\fi



%


\bibliographystyle{IEEEtran} 
\bibliography{mybib}

\end{document}